\documentclass[usenames,svgnames]{article}
\usepackage{ijcai25}
\usepackage{xcolor}
\definecolor{deepblue}{RGB}{0, 0, 139} 

\usepackage{times}
\usepackage{float}
\usepackage{soul}
\usepackage{url}
\usepackage[hidelinks]{hyperref}
\hypersetup{
    colorlinks=true,
    linkcolor=deepblue,      
    citecolor=deepblue,      
    filecolor=blue,      
    urlcolor=deepblue
}
\usepackage[utf8]{inputenc}
\usepackage[small]{caption}
\usepackage{graphicx}
\usepackage{amsmath}
\usepackage{amsthm}
\usepackage{booktabs}
\usepackage{algorithm}
\usepackage{algorithmic}
\usepackage[switch]{lineno}
\usepackage{tabularx}
\usepackage{booktabs}
\usepackage{pifont}
\usepackage{mydef}
\usepackage{booktabs}
\usepackage{tabularx}
\usepackage{framed}
\usepackage{bbding}
\usepackage{subfloat}
\usepackage{adjustbox}
\usepackage{multirow}
\usepackage{pifont}
\usepackage{natbib}

\newcommand{\paratitle}[1]{\vspace{0.8ex}\noindent 
\textbf{#1}}
\title{
How to Enable LLM with 3D Capacity? A Survey of Spatial Reasoning in LLM 
}

\author{
Jirong Zha$^{1*}$
\and
Yuxuan Fan$^{2*}$
\and
Xiao Yang$^{2}$
\and
Chen Gao$^{1 \dag}$
\and
Xinlei Chen$^{1 \dag}$ \\
\affiliations
$^1$Tsinghua University \\
$^2$The Hong Kong University of Science and Technology (Guang Zhou)\\
\emails
zhajirong23@mails.tsinghua.edu.cn,
\{yfan546, xyang856\}@connect.hkust-gz.edu.cn,
chgao96@gmail.com,
chen.xinlei@sz.tsinghua.edu.cn}

\begin{document}

\maketitle

\begin{abstract}
3D spatial understanding is essential in real-world applications such as robotics, autonomous vehicles, virtual reality, and medical imaging. Recently, Large Language Models (LLMs), having demonstrated remarkable success across various domains, have been leveraged to enhance 3D understanding tasks, showing potential to surpass traditional computer vision methods. In this survey, we present a comprehensive review of methods integrating LLMs with 3D spatial understanding. We propose a taxonomy that categorizes existing methods into three branches: image-based methods deriving 3D understanding from 2D visual data, point cloud-based methods working directly with 3D representations, and hybrid modality-based methods combining multiple data streams. We systematically review representative methods along these categories, covering data representations, architectural modifications, and training strategies that bridge textual and 3D modalities. Finally, we discuss current limitations, including dataset scarcity and computational challenges, while highlighting promising research directions in spatial perception, multi-modal fusion, and real-world applications. 
\end{abstract}

\section{Introduction}
Large Language Models (LLMs) have evolved from basic neural networks to advanced transformer models like BERT \citep{kenton2019bert} and GPT \citep{radford2018improving}, originally excelling at language tasks by learning from vast text datasets. Recent advancements, however, have extended these models beyond pure linguistic processing to encompass multimodal ability (In this paper, when we refer to LLMs, we specifically mean those that integrate multimodal functions). Their ability to capture complex patterns and relationships \citep{chen2024spatialvlm} now holds promise for spatial reasoning tasks \citep{ma2024llms}. By applying these enhanced models to challenges such as understanding 3D object relationships and spatial navigation, we open up new opportunities for advancing fields like robotics, computer vision, and augmented reality \citep{gao2024embodiedcity}.

\begin{figure}[t]
    \centering
    \includegraphics[width=1\linewidth]{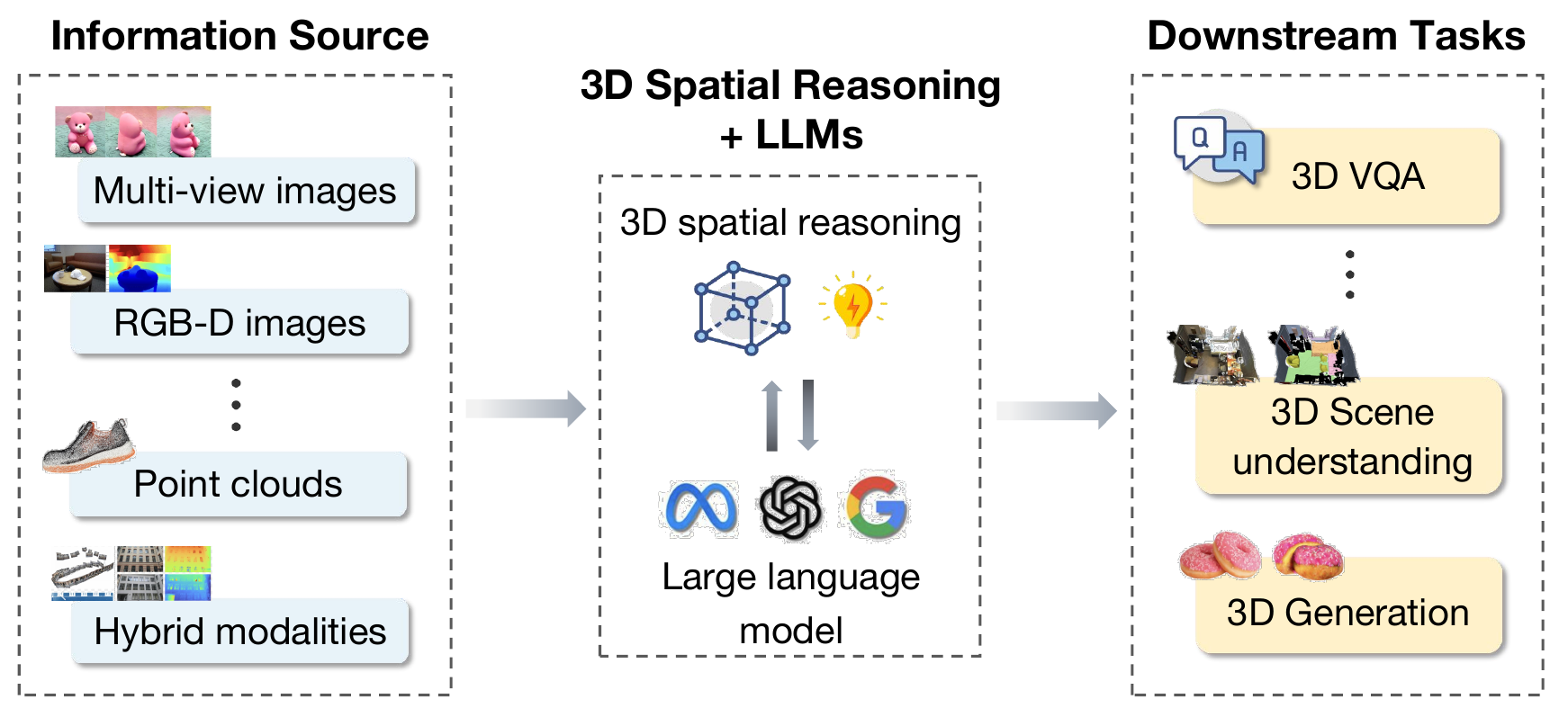}
    \caption{Large Language Models can acquire 3D spatial reasoning capabilities through various input sources including multi-view images, RGB-D images, point clouds, and hybrid modalities, enabling the processing and understanding of three-dimensional information.}
    \label{fig:overview}
\end{figure}

At the same time, 3D data and 3D modeling techniques have seen significant developments \citep{ma2024vision}, finding extensive applications in virtual and augmented reality, robotics, autonomous vehicles, gaming, medical imaging, and more. Unlike traditional two-dimensional images, 3D data provides a richer view of objects and environments, capturing essential spatial relationships and geometry. Such information is critical for tasks like scene reconstruction, object manipulation, and autonomous navigation, where merely text-based descriptions or 2D representations may fall short of conveying the necessary depth or spatial context.

\paratitle{LLMs help Spatial Understanding.} Bringing these two fields together—powerful language understanding from LLMs and the spatial realism of 3D data—offers the potential for highly capable, context-aware systems. From a linguistic perspective, real-world descriptions often reference physical arrangement, orientation, or manipulations of objects in space. Text alone can be imprecise or ambiguous about size, shape, or relative positioning unless one can integrate a robust spatial or visual understanding. Consequently, there is growing interest in enhancing LLMs with a “3D capacity” that enables them to interpret, reason, and even generate three-dimensional representations based on natural language prompts. Such an integrated approach opens up exciting prospects: robots that can follow language instructions more effectively by grounding their commands in 3D context, architects who quickly prototype 3D layouts from textual descriptions, game designers who generate immersive environments for narrative-based experiences, and many other creative applications yet to be envisioned.

\paratitle{Motivation.} Although LLMs have been increasingly applied in 3D-related tasks, and \cite{ma2024llms} provided a systematic overview of this field, the rapid advancement of this domain has led to numerous new developments in recent months, necessitating an up-to-date survey that captures these recent breakthroughs.
Integrating 3D capacity into LLMs faces several key challenges: (1) the scarcity of high-quality 3D datasets compared to abundant text corpora; (2) the fundamental mismatch between sequential text data and continuous 3D spatial structures, requiring specialized architectural adaptations; and (3) the intensive computational requirements for processing 3D data at scale. While early attempts at combining language and 3D have shown promise, current approaches often remain limited in scope, scalability, and generalization capability. Most existing solutions are domain-specific and lack the broad applicability characteristic of text-based LLMs.

\paratitle{Contribution.} The contributions of this work are summarized in the following three aspects: \textbf{(1)} \textit{A structured taxonomy}. We provide a timely and comprehensive survey that distinguishes itself from the systematic overview offered by \cite{ma2024llms} by presenting a novel perspective on LLM applications in 3D-related tasks: our work constructs a structured taxonomy that categorizes existing research into three primary groups (Figure \ref{fig:taxonomy_of_3dllm}) and offers a forward-looking analysis of the latest breakthroughs, thereby underscoring our unique contributions and the significance of our approach in advancing the field. \textbf{(2)} \textit{A comprehensive review}. Building on the proposed taxonomy, we systematically review the current research progress on LLMs for spatial reasoning tasks. \textbf{(3)} \textit{Future directions}. We highlight the remaining limitations of existing works and suggest potential directions for future research.

\section{Preliminary}
\tikzstyle{leaf}=[draw=hiddendraw,
    rounded corners,minimum height=1em,
    fill=mygreen!10,text opacity=1, align=center,
    fill opacity=.5,  text=black,align=left,font=\scriptsize,
    inner xsep=3pt,
    inner ysep=1pt,
    ]
\tikzstyle{middle}=[draw=hiddendraw,
    rounded corners,minimum height=1em,
    fill=output-white!40,text opacity=1, align=center,
    fill opacity=.5,  text=black,align=left,font=\scriptsize,
    inner xsep=3pt,
    inner ysep=1pt,
    ]
\begin{figure*}[ht]
\centering
\begin{forest}
  for tree={
  forked edges,
  grow=east,
  reversed=true,
  anchor=base west,
  parent anchor=east,
  child anchor=west,
  base=middle,
  font=\scriptsize,
  rectangle,
  line width=0.7pt,
  draw=output-black,
  rounded corners,align=left,
  minimum width=2em,
    s sep=5pt,
    inner xsep=3pt,
    inner ysep=1pt,
  },
  where level=1{text width=4.5em}{},
  where level=2{text width=6em,font=\scriptsize}{},
  where level=3{font=\scriptsize}{},
  where level=4{font=\scriptsize}{},
  where level=5{font=\scriptsize}{},
  [3D Spatial Reasoning in Large Language Model, middle,rotate=90,anchor=north,edge=output-black
    [\textbf{Image}-based, middle, edge=output-black,text width=6.4em
        [Multi-view Images as Input, middle, text width=8.3em, edge=output-black
            [LLaVA-3D \citep{zhu2024llava}{,} 
             Agent3D-Zero \citep{zhang2024agent3d}{,}
             \\ 
             ShapeLLM \citep{qi2024shapellm} 
             Scene-LLM \citep{fu2024scene}
             3d-LLM \citep{hong20233d}
             , leaf, text width=27.3em, edge=output-black]
        ]
        [RGB-D Images as Input, middle, text width=8.3em, edge=output-black
            [
            SpatialPIN \citep{SpatialPIN}{,} M3D-LaMed \citep{bai2024m3d}
            , leaf, text width=27.3em, edge=output-black]]
        [Monocular Images as Input, middle, text width=8.3em, edge=output-black
            [LLMI3D \citep{yang2024llmi3d}{,}
            Spatialvlm \citep{chen2024spatialvlm}
            , leaf, text width=27.3em, edge=output-black]
        ]
        [Medical Images as Input, middle, text width=8.3em, edge=output-black
            [M3D-LaMed \citep{bai2024m3d}{,} HILT \citep{liu2024benchmarking}{,} 3D-CT-GPT \citep{chen20243d}{,}\\ OpenMEDLab \citep{wang2024openmedlab}
            , leaf, text width=27.3em, edge=output-black]
        ]
    ]
    [\textbf{Point cloud}-based, middle, edge=output-black, text width=6.4em
        [Direct Alignment, middle, text width=8.3em, edge=output-black
            [PointLLM  \citep{xu2025pointllm}{,}
             Chat-Scene \citep{huang2024chat}{,}
             \\
             PointCLIP \citep{zhang2022pointclip}{,} 
             PointCLIPv2 \citep{zhu2023pointclip}
            , leaf, text width=27.2em, edge=output-black]
        ]
        [Step-by-step Alignment, middle, text width=8.3em, edge=output-black
            [GPT4Point \citep{qi2024gpt4point}{,} 
             MiniGPT-3D \citep{tang2024minigpt}{,}
             \\
             GreenPLM \citep{tang2024more}{,}
             Grounded 3D-LLM \citep{chen2024grounded}{,}
             \\
             Lidar-LLM\citep{yang2023lidar}{,}
            , leaf, text width=27.2em, edge=output-black]
        ]
        [Task-specific Alignment, middle, text width=8.3em, edge=output-black
            [3D-LLaVA  \citep{deng20253d}{,} 
             ScanReason \citep{zhu2024scanreason}{,} 
             SegPoint \citep{he2024segpoint}{,} 
             \\
             Kestrel \citep{fei2024kestrel}{,} 
             SIG3D \citep{man2024situational}{,} 
             Chat-3D  \citep{wang2023chat}{,}
             \\ 
             LL3DA \citep{chen2024ll3da}{,}
            , leaf, text width=27.2em, edge=output-black]
        ]
    ]
    [\textbf{Hybrid modality}-based, middle, edge=output-black, text width=7em
        [Tightly Coupled, middle, text width=8.3em, edge=output-black
            [Point-bind  \citep{guo2023point}{,} 
             JM3D \citep{ji2024jm3d}{,}
             Uni3D \citep{zhou2023uni3d}
             \\
             Uni3D-LLM \citep{liu2024uni3d}
            , leaf, text width=27.2em, edge=output-black]
        ]
        [Loosely Coupled, middle, text width=8.3em, edge=output-black
            [MultiPLY  \citep{hong2024multiply}{,} 
             UniPoint-LLM \citep{liupointmllm}
            , leaf, text width=27.2em, edge=output-black]
        ]
    ]
  ]
\end{forest} 
\caption{A Taxonomy of Models for Spatial Reasoning with LLMs: Image-based, Point Cloud-based, and Hybrid Modality-based Approaches and Their Subdivisions.}
\label{fig:taxonomy_of_3dllm}
\end{figure*}
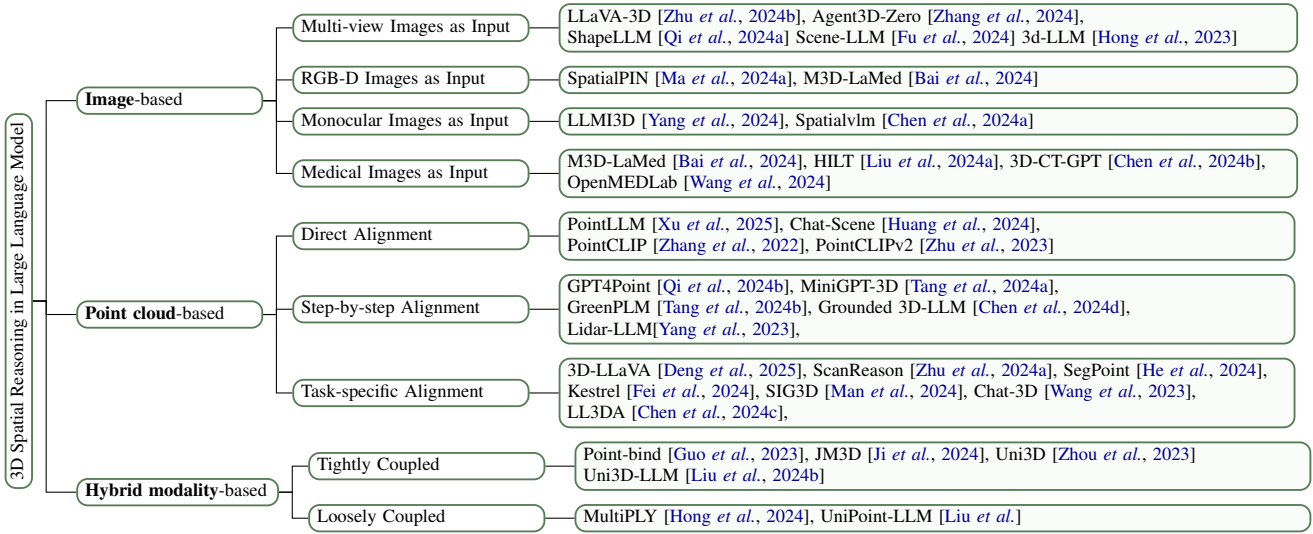
\subsection{Large Language Models}
Large Language Models (LLMs) have evolved from early word embeddings to context-aware models like BERT \citep{kenton2019bert}. Generative transformers such as GPT series \citep{radford2018improving}, have further advanced text generation and few-shot learning. However, these models often struggle with spatial reasoning due to their focus on textual patterns, prompting efforts to integrate external spatial knowledge \citep{fu2024scene}.

Vision-Language Models (VLMs) extend LLMs by aligning visual data with text. Early examples like CLIP \citep{radford2021learning} leverage co-attentional architectures and contrastive learning, while later models such as BLIP \citep{li2022blip} refine these techniques with larger datasets. Yet, most VLMs process only 2D data, limiting their ability to capture detailed 3D spatial configurations. Integrating 3D context via depth maps, point clouds, or voxels remains challenging, motivating ongoing research toward more robust spatial intelligence.

\subsection{3D Data Structures}
3D data has different structures, which are essential for understanding the three-dimensional world, and common methods include point clouds, voxel grids, polygonal meshes, neural fields, hybrid representations, and 3D Gaussian splatting. Point clouds represent shapes using discrete points, typically denoted as $$P = \left\{p_i \in \mathbb{R}^3 \mid i = 1,...,N\right\},$$
which are storage-efficient but lack surface topology. Voxel grids partition space into uniform cubes, with each voxel $V(i,j,k)$ storing occupancy or distance values, providing detailed structure at the expense of increased memory usage at higher resolutions. Polygonal meshes compactly encode complex geometries through a set of vertices $\{v_i\}$ and faces $\{F_j\}$, though their unstructured and non-differentiable nature poses challenges for integration with neural networks. Neural fields offer an implicit approach by modeling 3D shapes as continuous and differentiable functions, such as $$f_\theta : \mathbb{R}^3 \rightarrow (c,\sigma),$$ which maps spatial coordinates to color $c$ and density $\sigma$. Hybrid representations combine these neural fields with traditional volumetric methods (e.g., integrating $f_\theta$ with voxel grids) to achieve high-quality, real-time rendering. Meanwhile, 3D Gaussian splatting enhances point clouds by associating each point $p_i$ with a covariance matrix $\Sigma_i$ and color $c_i$, efficiently encoding radiance information for rendering. Each method has its unique strengths and trade-offs, making them suitable for different applications in 3D understanding and generation.

\subsection{Proposed taxonomy}
We propose a taxonomy that classifies 3D-LLM research into three main categories based on input modalities and integration strategies, as shown in Figure \ref{fig:overview}: Image-based spatial reasoning encompasses approaches that derive 3D understanding from 2D images. This includes multi-view methods that reconstruct 3D scenes, RGB-D images providing explicit depth information, monocular 3D perception inferring depth from single views, and medical imaging applications. While these approaches benefit from readily available image data and existing vision models, they may struggle with occlusions and viewpoint limitations. Point cloud-based spatial reasoning works directly with 3D point cloud data through three alignment strategies: (1) Direct alignment that immediately connects point features with language embeddings, (2) Step-by-step alignment that follows sequential stages to bridge modalities, and (3) Task-specific alignment customized for particular spatial reasoning requirements. These methods maintain geometric fidelity but face challenges in handling unstructured 3D data. Hybrid modality-based spatial reasoning combines multiple data streams through either tightly or loosely coupled architectures. Tightly coupled approaches integrate modalities through shared embeddings or end-to-end training, while loosely coupled methods maintain modular components with defined interfaces between them. This enables leveraging complementary strengths across modalities but increases architectural complexity.

This taxonomy provides a structured framework for understanding the diverse technical approaches in the field while highlighting the distinct challenges and trade-offs each branch must address. Figure \ref{fig:taxonomy_of_3dllm} presents a detailed breakdown of representative works in each category.


\section{Recent Advances of Spatial Reasoning in LLM}
\subsection{Image-based Spatial Reasoning}
\begin{figure}
    \centering
    \includegraphics[width=1\linewidth]{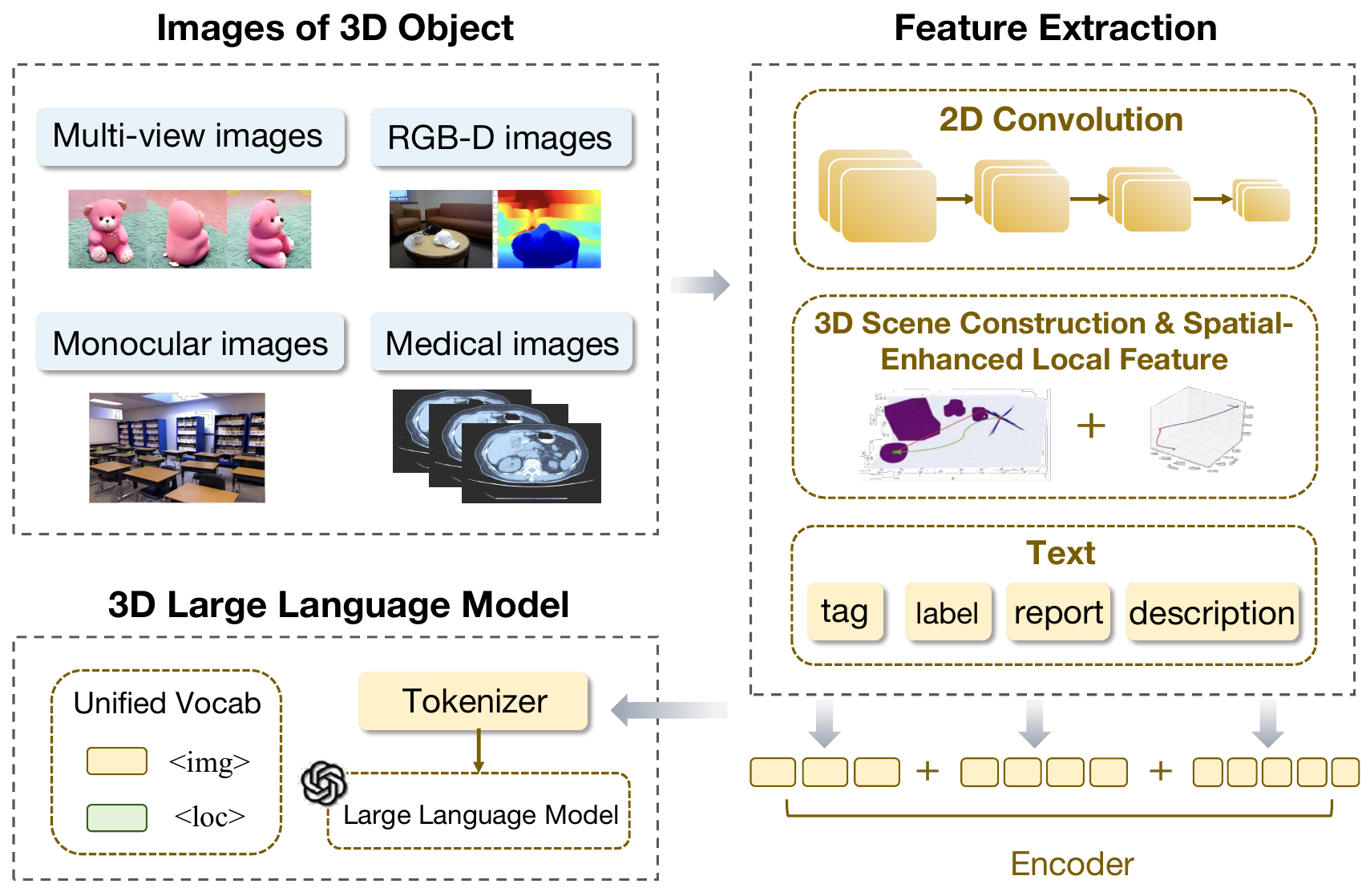}
    \caption{An overview of image-based approaches.}
    \label{fig:image-based}
\end{figure}

Image-based spatial reasoning methods can be categorized based on their input modalities: multi-view images, monocular images, RGB-D images, and 3D medical images shown in Figure \ref{fig:image-based}. Each modality offers unique advantages for enhancing 3D understanding in Large Language Models (LLMs). Multi-view images provide spatial data from different perspectives, monocular images extract 3D insights from a single view, RGB-D images incorporate depth information, and 3D medical images address domain-specific challenges in healthcare. These categories highlight the strengths and challenges of each approach in improving spatial reasoning capabilities.
\subsubsection{Multi-view Images as input}
Several studies explore multi-view images to enhance LLMs' spatial understanding. LLaVA-3D \cite{zhu2024llava} leverages multi-view images and 3D positional embeddings to create 3D Patches, achieving state-of-the-art 3D spatial understanding while maintaining 2D image understanding capabilities. Agent3D-Zero \cite{zhang2024agent3d} utilizes multiple images from different viewpoints, enabling VLMs to perform robust reasoning and understand spatial relationships, achieving zero-shot scene understanding. ShapeLLM \cite{qi2024shapellm} also uses multi-view image input, with robustness to occlusions. Scene-LLM \cite{fu2024scene} uses multi-view images to build 3D feature representations, incorporating scene-level and egocentric 3D information to support interactive planning. SpatialPIN \cite{SpatialPIN} enhances VLM's spatial reasoning by decomposing, understanding and reconstructing explicit 3D representations from multi-view images and generalizes to various 3D tasks. LLMI3D \cite{yang2024llmi3d} extracts spatially enhanced local features from high-resolution images using CNNs and a depth predictor and uses ViT to obtain tokens from low-resolution images. It employs a spatially enhanced cross-branch attention mechanism to effectively mine spatial local features of objects and uses geometric projection to handle.
Extracting multi-view features results in huge computational overhead and ignores the essential geometry and depth information. Additionally, plain texts often lead to ambiguities especially in cluttered and complex 3D environments\cite{chen2024ll3da}. ConceptGraphs \cite{gu2024conceptgraphs} proposes a graph-structured representation for 3D scenes that operates with an open vocabulary, which is developed by utilizing 2D foundation models and integrating their outputs into a 3D format through multiview association.
\subsubsection{Monocular Image as input}
LLMI3D \cite{yang2024llmi3d} uses a single 2D image for 3D perception, enhancing performance through spatial local feature mining, 3D query token decoding, and geometry-based 3D reasoning. It uses a depth predictor and CNN to extract spatial local features and uses learnable 3D query tokens for geometric coordinate regression. It combines black-box networks and white-box projection to address changes in camera focal lengths.
\subsubsection{RGB-D Image as Input}
Depth is estimated in SpatialPIN \cite{SpatialPIN} by ZoeDepth when finding field of view (FOV) through perspective fields and provided for 3D-scene understanding and reconstruction. M3D-LaMed \cite{bai2024m3d} pre-trains the 3D medical vision encoder with medical image slices along depth and introduces end-to-end tuning to integrate 3D information into LLM.

\subsubsection{3D Medical Image as input}
Unlike previous research focused on 2D medical images, integrating multi-modal other information such as textual descriptions, M3D-LaMed \cite{bai2024m3d} is specifically designed for 3D CT images by analyzing spatial features. It demonstrates excellent performance across multiple tasks, including image-text retrieval, report generation, visual question answering, localization, and segmentation. In order to generate radiology reports automatically, a brand-new framework  \cite{liu2024benchmarking} is proposed to employs low-resolution (LR) visual tokens as queries to extract information from high-resolution (HR) tokens, ensuring that detailed information is retained across HR volumes while minimizing computational costs by processing only the HR-informed LR visual queries. 3D-CT-GPT \cite{chen20243d}, based medical visual language model, is tailored for the generation of radiology reports from 3D CT scans, with a focus on chest CTs. OpenMEDLab \cite{wang2024openmedlab} comprises and publishes a variety of medical foundation models to process multi-modal medical data including Color Fundus Photography (CFP), Optical Coherence Tomography (OCT), endoscopy videos, CT\&MR volumes and other pathology images.

\subsubsection{Discussion}
Image-based spatial reasoning methods offer significant advantages, such as easy data acquisition and integration with pre-trained 2D models. Multi-view images provide rich spatial information, while depth estimation enhances scene understanding. However, challenges remain, including limited depth from single views, scale uncertainty, occlusion, and viewpoint dependency. These methods also face issues with visual hallucinations, generalization to novel scenes, and high computational costs. Future research should focus on improving multi-view integration and depth estimation to address these limitations.
\subsection{Recent Advances of Point Cloud-based Spatial Reasoning}
\begin{figure}
    \centering
    \includegraphics[width=1\linewidth]{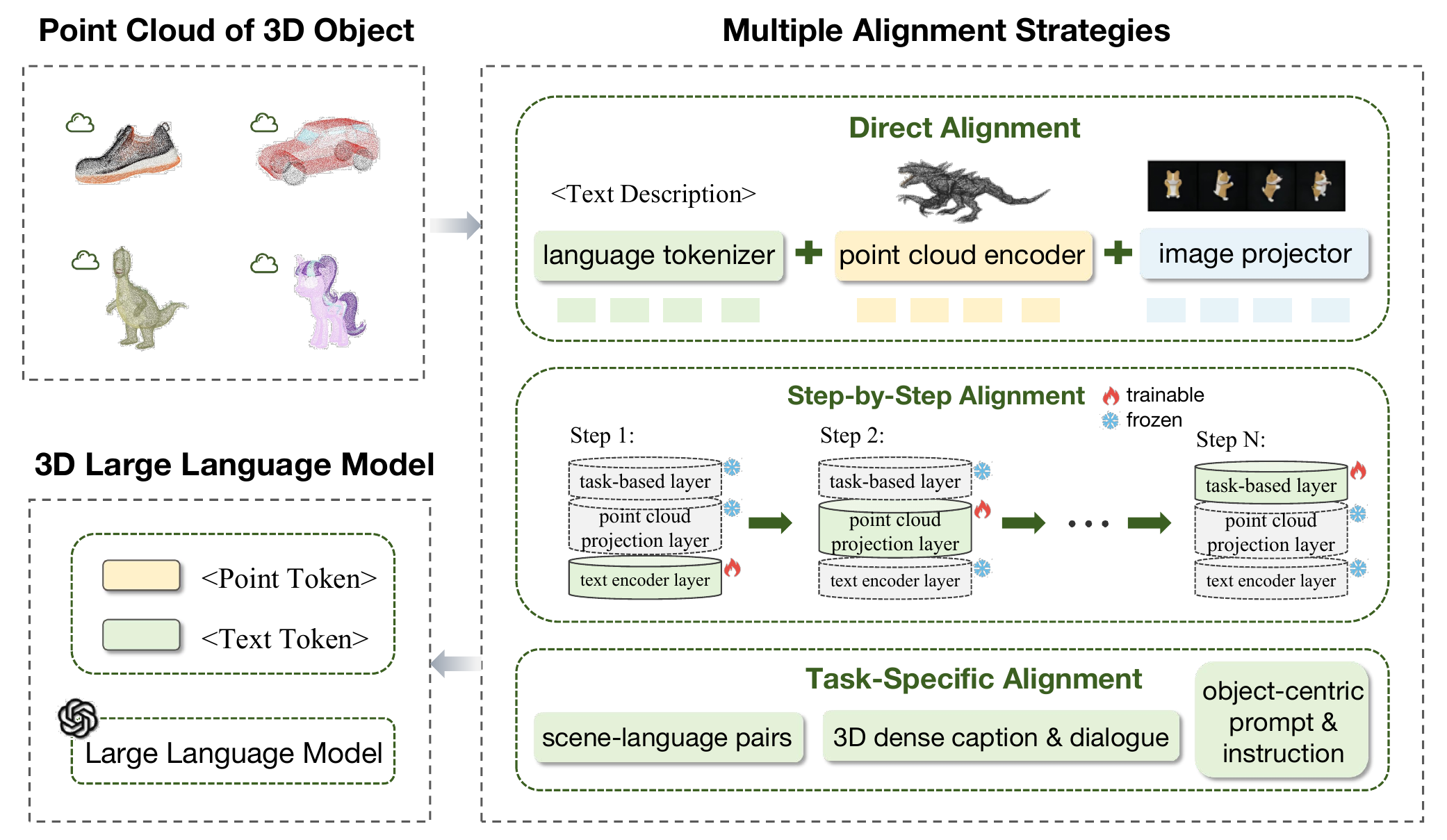}
    \caption{An overview of point cloud-based approaches.}
    \label{fig:point}
\end{figure}

As shown in Figure \ref{fig:point}, point cloud-based spatial reasoning has advanced significantly in recent years, employing three main alignment methods: Direct, Step-by-step, and Task-specific Alignment. These methods are essential for integrating point cloud data with language models to enable effective spatial reasoning. Direct Alignment establishes immediate connections between point cloud features and language model embeddings, while Step-by-step Alignment follows a sequential process through multiple stages. Task-specific Alignment is customized for particular spatial reasoning requirements. The choice of method depends on specific application needs and constraints. 
\subsubsection{Direct Alignment}
Direct alignment methods create direct connections between point cloud data and language models. PointCLIP \citep{zhang2022pointclip} was a pioneer, projecting point clouds into multi-view depth maps and using CLIP’s pre-trained visual encoder for feature extraction, which was then aligned with textual features through a hand-crafted template. This approach showed promising results in zero-shot and few-shot classification tasks by transferring 2D knowledge to the 3D domain. PointCLIP V2 \citep{zhu2023pointclip} improved the projection quality with a realistic projection module and used GPT-3 for generating 3D-specific text descriptions, achieving better performance in zero-shot classification, part segmentation, and object detection. Chat-Scene \citep{huang2024chat} introduced object identifiers to facilitate object referencing during user-assistant interactions, representing scenes through object-centric embeddings. PointLLM \citep{xu2025pointllm} advanced the field by integrating a point cloud encoder with a powerful LLM, effectively fusing geometric, appearance, and linguistic information, and overcoming data scarcity with automated generation. These methods demonstrate the potential for effective 3D point cloud understanding through language models, enabling improved spatial reasoning and human-AI interaction.
\subsubsection{Step-by-step Alignment}
Step-by-step alignment has gained popularity in integrating point cloud features with language models. Notable approaches include GPT4Point \citep{qi2024gpt4point}, which uses a Bert-based Point-QFormer for point-text feature alignment, followed by object generation. Grounded 3D-LLMs \citep{chen2024grounded} first aligns 3D scene embeddings with textual descriptions via contrastive pre-training, then fine-tunes with referent tokens. LiDAR-LLMs \citep{yang2023lidar} employ a three-stage process: cross-modal alignment, object-centric learning, and high-level instruction fine-tuning. MiniGPT-3D \citep{tang2024minigpt} follows a four-stage strategy, from point cloud projection to advanced model enhancements using Mixture of Query Experts. GreenPLM \citep{tang2024more} uses a three-stage method that aligns a text encoder with an LLM using large text data, followed by point-LLM alignment with 3D data. These step-by-step approaches highlight the gradual improvement of spatial reasoning in 3D contexts, offering valuable insights for future research.
\subsubsection{Task-specific Alignment}
Task-specific alignment customizes models for specific spatial reasoning tasks to improve performance and generalization. SceneVerse \citep{jia2024sceneverse} introduces a large 3D vision-language dataset and Grounded Pre-training for Scenes (GPS), using multi-level contrastive alignment for unified scene-text alignment, achieving state-of-the-art results in tasks like 3D visual grounding and question answering. LL3DA \citep{chen2024ll3da} presents a dialogue system that integrates textual instructions and visual interactions, excelling in complex 3D environments. Chat-3D \citep{wang2023chat} proposes a three-stage training scheme to align 3D scene representations with language models, capturing spatial relations with limited data. VisProg \citep{yuan2024visual} introduces visual programming for zero-shot open-vocabulary 3D grounding, leveraging LLMs to generate and execute programmatic representations. These task-specific approaches highlight the importance of adapting models to complex spatial relationships, enabling robust performance even with limited data or zero-shot tasks.
\begin{table*}[t]
\resizebox{\linewidth}{!}{
\begin{tabular}{lllccclc}
\toprule
   & \textbf{Model} & \textbf{Data Source} & \textbf{Alignment Type} & \textbf{Pre-training}   & \textbf{Fine-tuning} & \textbf{Task} & \textbf{Code} \\ \midrule
 
\multirow{12}{*}{\rotatebox{90}{Image - based}}    
& LLaVA-3D \citep{zhu2024llava} & Multi-view Images & -  & \ding{51} & \ding{51}  & 3D VQA, 3D Scene Understanding & \href{https://github.com/ZCMax/LLaVA-3D}{code} \\
& Agent3D-Zero \citep{zhang2024agent3d} & Multi-view Images & -  & \ding{51} & \ding{55}  & 3D VQA, 3D Semantic Segmentation & \ding{55} \\
& ShapeLLM \citep{qi2024shapellm} & Multi-view Images & -  & \ding{51} & \ding{51}  & 3D Object Classification, 3D Scene Captioning & \href{https://github.com/qizekun/ShapeLLM}{code} \\
& Scene-LLM \citep{fu2024scene} & Multi-view Images & -  & \ding{51} & \ding{51}  & 3D VQA, Dense Captioning & \ding{55} \\
& SpatialPIN \citep{SpatialPIN} & RGB-D Images & -  & \ding{51} & \ding{55}  & 3D Motion Planning, Task Video Generation & \ding{55} \\
& LLMI3D \citep{yang2024llmi3d} & Monocular Images & -  & \ding{51} & \ding{51}  & 3D Grounding, 3D VQA & \ding{55} \\
& Spatialvlm \citep{chen2024spatialvlm} & Monocular Images & -  & \ding{51} & \ding{51}  & Dense Reward Annotator, Spatial Data Generation  & \href{https://github.com/jiayuww/SpatialEval}{code} \\
& M3D-LaMed \citep{bai2024m3d} & Medical Images & -  & \ding{51} & \ding{51}  & 3D VQA, 3D VLP & \href{https://github.com/BAAI-DCAI/M3D}{code} \\
& HILT \citep{liu2024benchmarking} & Medical Images & -  & \ding{51} & \ding{51}  & 3DHRG & \ding{55} \\
& 3D-CT-GPT \citep{chen20243d} & Medical Images & -  & \ding{51} & \ding{51}  & Radiology Report Generation, 3D VQA & \ding{55} \\
& OpenMEDLab \citep{wang2024openmedlab} & Medical Images & -  & \ding{51} & \ding{51}  & Medical Imaging & \href{https://github.com/openmedlab}{code} \\
                                   \midrule
\multirow{18}{*}{\rotatebox{90}{Point Cloud - based}}  
&  PointLLM \citep{xu2025pointllm} & Point Cloud & Direct Alignment  & \ding{51} & \ding{51}  & 3D Object Classification, 3D Object Captioning & \href{https://github.com/OpenRobotLab/PointLLM}{code} \\ 
&  Chat-Scene \citep{huang2024chat} & Point Cloud & Direct Alignment  & \ding{51} & \ding{51}  & 3D Visual Grounding, 3D Scene Captioning & \href{https://github.com/ZzZZCHS/Chat-Scene}{code} \\ 
&  PointCLIP \citep{zhang2022pointclip} & Point Cloud & Direct Alignment  & \ding{51} & \ding{51}  & 3D Point Cloud Classification & \href{https://github.com/ZrrSkywalker/PointCLIP}{code} \\
&  PointCLIPv2 \citep{zhu2023pointclip} & Point Cloud & Direct Alignment  & \ding{51} & \ding{51}  & 3D Point Cloud Classification & \href{https://github.com/yangyangyang127/PointCLIP_V2}{code} \\
&  GPT4Point \citep{qi2024gpt4point} & Point Cloud & Step-by-step Alignment  & \ding{51} & \ding{51}  & 3D Object Understanding & \href{https://github.com/Pointcept/GPT4Point}{code} \\
&  MiniGPT-3D \citep{tang2024minigpt} & Point Cloud & Step-by-step Alignment  & \ding{51} & \ding{51}  & 3D Object Classification, 3D Object Captioning & \href{https://github.com/tangyuan96/minigpt-3d}{code} \\
&  GreenPLM \citep{tang2024more} & Point Cloud & Step-by-step Alignment  & \ding{51} & \ding{51}  & 3D Object Classification & \href{https://github.com/TangYuan96/GreenPLM}{code} \\
&  Grounded 3D-LLM \citep{chen2024grounded} & Point Cloud & Step-by-step Alignment  & \ding{51} & \ding{51}  & 3D Object Detection, 3D VQA & \href{https://github.com/OpenRobotLab/Grounded_3D-LLM}{code} \\
&  Lidar-LLM \citep{yang2023lidar} & Point Cloud & Step-by-step Alignment  & \ding{51} & \ding{51}  & 3D Captioning, 3D Grounding & \href{https://github.com/Yangsenqiao/LiDAR-LLM/tree/main}{code} \\
&  3D-LLaVA \citep{deng20253d} & Point Cloud & Task-specific Alignment  & \ding{51} & \ding{51}  & 3D VQA, 3D Captioning & \href{https://github.com/ZCMax/LLaVA-3D}{code} \\
&  ScanReason \citep{zhu2024scanreason} & Point Cloud & Task-specific Alignment  & \ding{51} & \ding{51}  & 3D Reasoning Grounding & \href{https://github.com/ZCMax/ScanReason}{code} \\
&  SegPoint \citep{he2024segpoint} & Point Cloud & Task-specific Alignment  & \ding{51} & \ding{51}  & 3D Instruction Segmentation & \ding{55} \\
&  Kestrel \citep{fei2024kestrel} & Point Cloud & Task-specific Alignment  & \ding{51} & \ding{51}  & Part-Aware Point Grounding & \ding{55} \\
&  SIG3D \citep{man2024situational} & Point Cloud & Task-specific Alignment  & \ding{51} & \ding{51}  & Situation Estimation & \href{https://github.com/YunzeMan/Situation3D}{code} \\
&  Chat-3D  \citep{wang2023chat} & Point Cloud & Task-specific Alignment  & \ding{51} & \ding{51}  & 3D VQA & \href{https://github.com/Chat-3D/Chat-3D}{code} \\
&  LL3DA \citep{chen2024ll3da} & Point Cloud & Task-specific Alignment  & \ding{51} & \ding{51}  & 3D Dense Captioning & \href{https://github.com/Open3DA/LL3DA}{code} \\
 \midrule
\multirow{6}{*}{\rotatebox{90}{Hybrid - based}} 
& Point-bind  \citep{guo2023point} & Point cloud, Image & Tightly Coupled  & \ding{51} & \ding{51}  & 3D Cross-modal Retrieval, Any-to-3D Generation & \href{https://github.com/ZiyuGuo99/Point-Bind_Point-LLM/}{code} \\
& JM3D \citep{ji2024jm3d} & Point cloud, Image & Tightly Coupled  & \ding{51} & \ding{51}  & Image-3D Retrieval, 3D Part Segmentation & \href{https://github.com/Mr-Neko/JM3D}{code} \\
& Uni3D \citep{zhou2023uni3d} & Point cloud, Image & Tightly Coupled  & \ding{51} & \ding{51}  & Zero-shot Shape Classification & \href{https://github.com/baaivision/Uni3D}{code} \\
& Uni3D-LLM \citep{liu2024uni3d} & Point cloud, Image & Tightly Coupled  & \ding{51} & \ding{51}  & 3D VQA & \ding{55} \\
& MultiPLY  \citep{hong2024multiply} & Point cloud, Image & Loosely Coupled  & \ding{51} & \ding{51}  & Object retrieval & \href{https://github.com/UMass-Embodied-AGI/MultiPLY}{code} \\ 
& UniPoint-LLM \citep{liupointmllm} & Point cloud, Image & Loosely Coupled  & \ding{51} & \ding{51}  & 3D generation, 3D VQA & \ding{55} \\
                      \bottomrule
\end{tabular}}
\caption{Taxonomy of Large Language Models with spatial reasoning capability. This table presents a comprehensive comparison of various 3D vision-language models categorized by their input modalities (image-based, point cloud-based, and hybrid-based), showing their data sources, alignment types, training strategies (pre-training and fine-tuning), primary tasks, and code availability. The models are organized into three main categories based on their input type: image-based models, point cloud-based models, and hybrid models that utilize both modalities.}
\label{tab:summarizations}
\end{table*}

\subsubsection{Discussion}
The three alignment approaches—Direct, Step-by-step, and Task-specific—each have distinct strengths and challenges. Direct alignment offers efficiency and quick results but struggles with complex spatial relationships. Step-by-step alignment improves feature integration at the cost of higher computational resources and training time. Task-specific alignment excels in specialized tasks but may lack broader applicability.

\subsection{Hybrid Modality-based Spatial Reasoning}
\begin{figure}[t]
    \centering
    \includegraphics[width=1\linewidth]{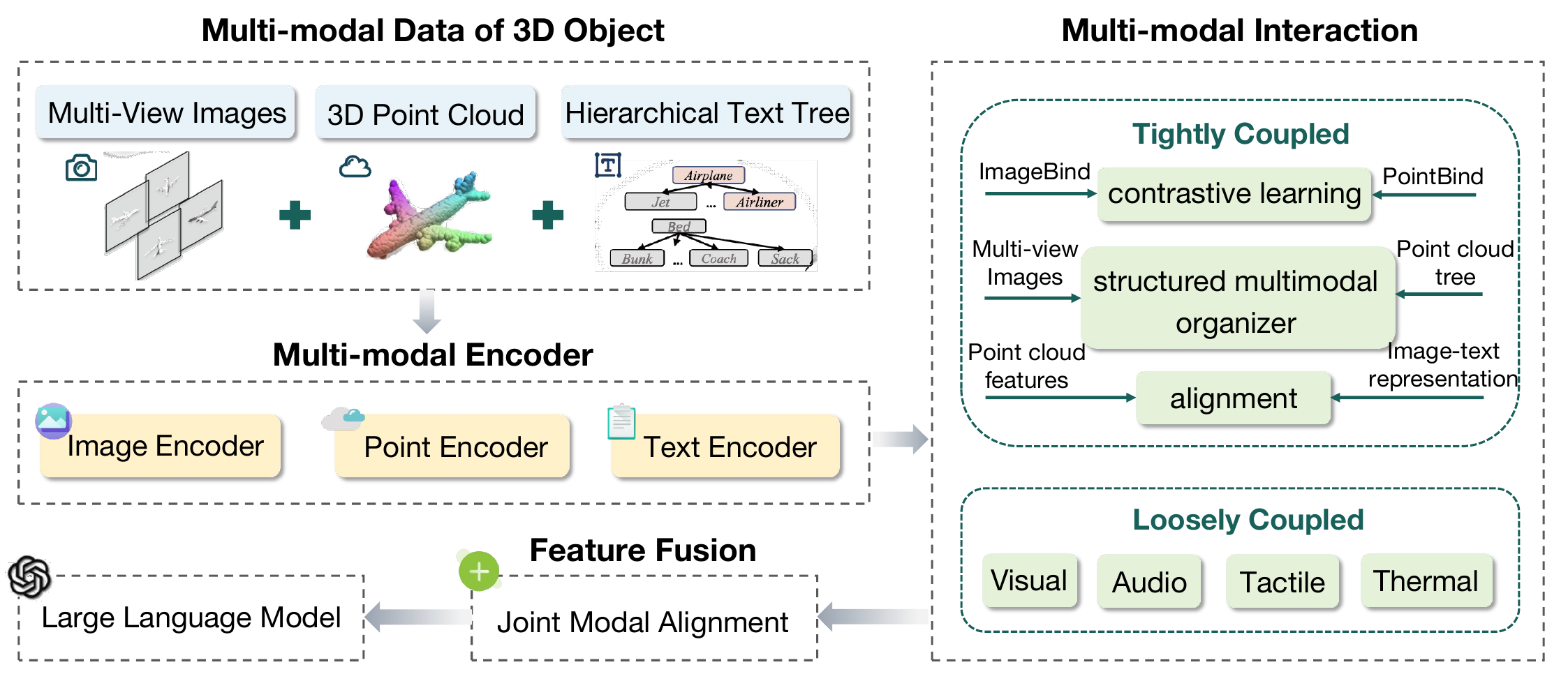}
    \caption{An overview of hybrid modality-based approaches.}
    \label{fig:hybrid}
\end{figure}

Hybrid modality-based spatial reasoning integrates point clouds, images, and LLMs through Tightly Coupled and Loosely Coupled approaches, as shown in Figure \ref{fig:hybrid}. The Tightly Coupled approach fosters close integration, enabling seamless interaction and high performance, while the Loosely Coupled approach promotes modularity, allowing independent operation of components for greater scalability and flexibility at the cost of reduced real-time interaction.
\subsubsection{Tightly Coupled}
Several recent works have explored tightly integrated approaches for spatial reasoning across point clouds, images and language modalities: Point-Bind \citep{guo2023point} proposes a joint embedding space to align point clouds with images and text through contrastive learning. It leverages ImageBind to construct unified representations that enable tasks like zero-shot classification, open-world understanding and multi-modal generation. The tight coupling allows Point-Bind to reason about point clouds using both visual and linguistic cues. JM3D \citep{ji2024jm3d} introduces a Structured Multimodal Organizer that tightly fuses multi-view images and hierarchical text trees with point clouds. This coupled architecture enables detailed spatial understanding by leveraging complementary information across modalities. The Joint Multi-modal Alignment further enhances the synergistic relationships between visual and linguistic features. Uni3D \citep{zhou2023uni3d} employs a unified transformer architecture that directly aligns point cloud features with image-text representations. By tightly coupling the modalities through end-to-end training, it achieves strong performance on tasks like zero-shot classification and open-world understanding. The shared backbone enables efficient scaling to billion-parameter models. Uni3D-LLM \citep{liu2024uni3d} extends this tight coupling to LLMs through an LLM-to-Generator mapping block. This enables unified perception, generation and editing of point clouds guided by natural language. The tight integration allows leveraging rich semantic knowledge from LLMs while maintaining high-quality 3D understanding.

\subsubsection{Loosely Coupled}
Loosely coupled approaches maintain greater independence between different modalities while still enabling interaction through well-defined interfaces. MultiPLY \citep{hong2024multiply} proposes a multisensory embodied LLM that handles multiple input modalities (visual, audio, tactile, thermal) through separate encoders. The modalities are processed independently and communicate through action tokens and state tokens. This decoupled design allows the system to process each modality with specialized encoders optimized for that data type, while enabling scalability and modularity in the system architecture. Similarly, UniPoint-LLM \citep{liupointmllm} introduces a Multimodal Universal Token Space (MUTS) that loosely connects point clouds and images through independent encoders and a shared mapping layer. This modular design allows easy integration of new modalities and simplified training by only requiring alignment between new modalities and text, rather than pairwise alignment between all modalities. The main benefits of loosely coupled architectures include greater modularity and flexibility in system design, easier integration of new modalities, and independent scaling of different components. However, this approach may result in less optimal joint representation learning, reduced real-time interaction capabilities, and potential information loss between modalities compared to tightly coupled approaches.

\subsubsection{Discussion}
The choice between tightly and loosely coupled approaches presents important tradeoffs in multimodal spatial reasoning systems. Tightly coupled approaches like Point-Bind and JM3D offer stronger joint representation learning and real-time interaction capabilities through end-to-end training and shared feature spaces. This makes them particularly suitable for applications requiring detailed spatial understanding and precise control. However, they can be more complex to train and scale, and adding new modalities may require significant architectural changes. In contrast, loosely coupled approaches like MultiPLY and UniPoint-LLM provide greater modularity and flexibility, making them easier to extend and maintain. They allow independent optimization of different components and simplified training procedures, but may sacrifice some performance in tasks requiring fine-grained cross-modal understanding. The optimal choice ultimately depends on specific application requirements - tightly coupled architectures may be preferred for specialized high-performance systems, while loosely coupled designs better suit general-purpose platforms prioritizing extensibility and maintainability. Future work may explore hybrid approaches that combine the benefits of both paradigms, potentially using adaptive coupling mechanisms that adjust based on task demands.

\section{Applications}
A key research focus leverages LLMs to enhance robotic embodied intelligence, enabling machines to interpret natural language commands for real-world tasks. This includes robotic control, navigation, and manipulation, where LLMs parse instructions, generate action plans, and adapt to dynamic environments—for instance, guiding robots to locate objects in cluttered spaces using text-based prompts.

\paratitle{3D Scene Understanding.}
Advanced 3D scene analysis integrates multimodal data (e.g., images, point clouds, text) for tasks like open-vocabulary segmentation, semantic mapping, and spatial reasoning. Central to this is 3D visual question answering (3D-VQA), requiring models to interpret queries about object attributes, spatial relationships, or contextual roles within scenes. Context-aware systems further account for user perspectives to deliver precise responses.

\paratitle{Cross-Domain Applications.}
In healthcare, LLMs analyze volumetric medical scans (e.g., CT) for lesion detection and automated diagnostics. Autonomous driving systems utilize 3D-capable LLMs to interpret traffic scenes, aiding object detection \citep{zha2023privacy, zha2024diffusion} and path planning. Design-oriented applications include generating indoor layouts from textual requirements, while educational tools employ interactive 3D environments to teach spatial concepts.

\section{Challenges and Future Directions}
Table \ref{tab:summarizations} summarizes the models that leverage LLMs to assist graph-related tasks according to the proposed taxonomy.
Based on the above review and analysis, we believe that there is still much space for further enhancement in this field.
Recent advances in integrating LLMs with three-dimensional (3D) data have demonstrated considerable promise. However, numerous challenges must still be overcome to realize robust and practical 3D-aware LLMs. Below, we summarize these obstacles and then outline potential pathways to address them, highlighting key research directions for the future.
\subsection{Challenges}
\paratitle{Weak Spatial Reasoning and Representation.}  Multimodal LLMs (MLLMs) exhibit limited acuity in 3D spatial understanding, struggling with fine-grained relationships (e.g., front/back distinctions, occluded object localization) and precise geometric outputs (distances, angles). These issues stem partly from mismatches between unstructured point clouds and sequence-based LLM architectures, where high-dimensional 3D data incur prohibitive token counts or oversimplified encodings.  

\paratitle{Data and Evaluation Gaps.}  Progress in 3D-aware LLMs is hindered by the scarcity of high-quality 3D-text paired datasets. Unlike the abundant resources for 2D images and video, the 3D domain lacks standardized, richly annotated datasets crucial for training robust models. Existing benchmarks focus mainly on discriminative tasks like classification and retrieval—emphasizing category differentiation rather than generating rich, descriptive 3D scene outputs. Consequently, evaluations often rely on subjective metrics (e.g., human or GPT-based judgments) that can lack consistency. Advancing the field requires developing objective, comprehensive benchmarks that assess both open-vocabulary generation and the spatial plausibility of descriptions relative to the underlying 3D structure. 

\paratitle{Multimodal Integration and Generalization.} Fusing 3D data (e.g., point clouds) with other modalities like 2D imagery, audio, or text poses significant challenges due to their distinct structural characteristics. The conversion and alignment of high-dimensional 3D data with lower-dimensional representations can lead to a loss of intricate details, diluting the original 3D richness. Moreover, current models often struggle with open-vocabulary recognition, limiting their ability to identify or describe objects outside of their training data—especially when encountering unseen scenes or novel objects. This difficulty is further compounded by the variability of natural language, from colloquial expressions to domain-specific terminology, and by noisy inputs. Thus, more sophisticated multimodal integration techniques and generalization strategies are needed to preserve geometric fidelity while accommodating diverse, unpredictable inputs.

\paratitle{Complex Task Definition.}  While 3D-aware LLMs excel in controlled settings, they lack frameworks for nuanced language-context inference in dynamic environments. Task decomposition and scalable encoding methods are needed to balance geometric fidelity with computational tractability, particularly for interactive applications requiring real-time spatial reasoning.

\subsection{Future Directions}
\paratitle{Enhancing 3D Perception and Representations.}  
Addressing spatial reasoning gaps requires richer 3D-text datasets (e.g., from robotics, gaming, autonomous driving) and model architectures that encode geometric relationships. Multi-view data and robust depth cues can improve orientation, distance, and occlusion estimation. Compact 3D tokens and refined encoding/decoding methods may bridge unstructured point clouds with sequence-based models, enabling fine-grained spatial understanding and generation.  

\paratitle{Multi-Modal Fusion and Instruction Understanding.}  
Tighter integration of modalities (point clouds, images, text, audio) via unified latent spaces or attention mechanisms could preserve subtle geometric and semantic details. Enhanced instruction processing—including hierarchical task decomposition, contextual interpretation, and robustness to dialects/terminology—would improve compositional reasoning in 3D environments and broaden real-world applicability. Furthermore, by leveraging these integrated representations, models can more adeptly adapt to complex instructions and novel scenarios, ultimately paving the way for more robust and versatile 3D reasoning systems. 

\paratitle{Cross-Scene Generalization and Robust Evaluation.}  
Open-vocabulary 3D understanding demands large-scale pretraining on diverse scenes and transfer/lifelong learning paradigms for adapting to novel objects or environments. This understanding extends beyond predefined categories to generalize to unseen objects and scenes. For instance, models need to comprehend ``an old rocking chair" even if this specific type of chair never appeared in the training data.  

\paratitle{Expanding Applications for Autonomous Systems.}  
3D-aware LLMs hold potential in robotics (navigation, manipulation), medical imaging (lesion detection), architectural design, and interactive education. Future systems may integrate environmental constraints, user perspectives, and object affordances for autonomous planning and decision-making in dynamic 3D contexts.

Collectively, these challenges and potential directions underscore the field’s rapid evolution and its equally significant open questions. Moving forward, more robust 3D-specific data resources, better model architectures, and more refined evaluation protocols will be essential to unlock the full potential of LLMs in three-dimensional settings—and ultimately bring intelligent, multimodal understanding closer to real-world deployment.

\section{Conclusion}
The integration of LLMs with 3D data is a dynamic research area. This survey categorized 3D-LLM research into image-based, point cloud-based, and hybrid modality-based spatial reasoning. It reviewed state-of-the-art methods, their applications in multiple fields, and associated challenges. Notably, image-based methods have data-related advantages but face issues like depth information shortage. Point cloud-based methods offer precise 3D details but encounter data-handling difficulties. Hybrid methods combine strengths yet struggle with data alignment. Applications are diverse, but challenges such as weak spatial perception, data scarcity, and evaluation problems exist. Future research should focus on enhancing 3D perception, improving multi-modal fusion, expanding generalization, developing evaluation metrics, enhancing instruction understanding, optimizing 3D representations, and exploring continuous learning. By addressing these, we can unlock the full potential of 3D-aware LLMs for real-world deployment and industry advancement. 






\bibliographystyle{named}
\bibliography{ijcai25}

\end{document}